\documentclass[hidelinks]{article}
\usepackage[utf8]{inputenc}
\usepackage[T1]{fontenc}
\usepackage{lmodern}
\usepackage{fix-cm}
\usepackage{microtype}

\usepackage{amsmath,amsfonts,amssymb,amsthm}
\usepackage{mathtools}
\usepackage{xcolor}
\usepackage{authblk}
\usepackage{graphicx}
\usepackage{booktabs}
\usepackage[section]{placeins}
\usepackage[super]{nth}
\usepackage{icomma}
\usepackage{pgfplots}
\pgfplotsset{width=7cm,compat=1.8}

\newcommand{\pvp}{P_{\mathit{TP}}}
\newcommand{\pvn}{P_{\mathit{TN}}}
\newcommand{\pfp}{P_{\mathit{FP}}}
\newcommand{\pfn}{P_{\mathit{FN}}}
\newcommand{\lfn}{L_{\mathit{FN}}}
\newcommand{\lfp}{L_{\mathit{FP}}}
\newcommand{\ex}[1]{\mathbb{E}[#1]}

\usepackage[authoryear]{natbib} 
\usepackage{hyperref}

\hypersetup{colorlinks=false}

\usepackage{caption}
\usepackage{subcaption}

\title{The Economic Implications of Large Language Model Selection on Earnings and Return on Investment: A Decision Theoretic Model}
\author[1]{Geraldo Xexéo}
\author[2]{Filipe Braida}
\author[1,3]{Marcus Parreiras}
\author[1]{Paulo Xavier}
\affil[1]{Programa de Engenharia de Sistemas e Computação -- COPPE \\
Universidade Federal do Rio de Janeiro, Brasil}
\affil[2]{Departamento de Ciência da Computação \\
Universidade Federal Rural do Rio de Janeiro}
\affil[3]{Coordenadoria de Engenharia de Produção - COENP \\ CEFET/RJ, Unidade Nova Iguaçu}

\date{May 27, 2024}

\begin{document}

\maketitle

\begin{abstract}
Selecting language models in business contexts requires a careful analysis of the final financial benefits of the investment.
However, the emphasis of academia and industry analysis of LLM is solely on performance.
This work introduces a framework to evaluate LLMs, focusing on the earnings and return on investment aspects that should be taken into account in business decision making.
We use a decision-theoretic approach to compare the financial impact of different LLMs, considering variables such as the cost per token, the probability of success in the specific task, and the gain and losses associated with LLMs use.
The study reveals how the superior accuracy of more expensive models can, under certain conditions, justify a greater investment through more significant earnings but not necessarily a larger RoI.
This article provides a framework for companies looking to optimize their technology choices, ensuring that investment in cutting-edge technology aligns with strategic financial objectives. 
In addition, we discuss how changes in operational variables influence the economics of using LLMs, offering practical insights for enterprise settings, finding that the predicted gain and loss and the different probabilities of success and failure are the variables that most impact the sensitivity of the models.
\end{abstract}

\section{Introduction}

This article proposes different approaches to large language model (LLM) evaluation by analyzing the financial impact that the adoption of these technologies can have on a business operation.
Although the vast majority of work only discusses the performance of LLMs in a set of tasks~\citep{chang2023survey}, here it is assumed that the process of selecting a Language Model (LLM) for specific tasks within a business context must go beyond performance assessment, looking also at operational aspects and taking into account the expected earnings and return on investment (RoI).

Our motivation is the fact that the selection of an appropriate LLM has become a strategic decision for companies looking to improve and optimize operations and maximize RoI~\citep{Gupta2024}.
With the continuous evolution of artificial intelligence technologies, LLMs offer innovative solutions for a variety of business applications, from virtual assistants to complex data analysis systems.
However, choosing between the different models available can be challenging considering variable costs and impacts on the performance of business tasks.
Furthermore, the constant evolution of the state-of-the-art, for different performance and cost levels, shows the need to always question the choice of LLM to be used at a certain moment, which requires constant evaluation of new models~\citep{shekhar2024optimizing}.

The paper presents different models to analyze the impacts of adopting different LLM based on their impact on earnings and return on investment. 
Each model supposes a scenario of business operations that includes a business task based on the execution of an LLM task. 
For example, LLMs can be used in a recommender system associated to an online sale~\citep{zhao2024recommender}, where it will propose a product to be acquired by the client. The business task is to actually sell the recommended system, while the LLM taks is find a correct recommendation. 
Usually, both tasks can be trained, fine-tuned, or tested using an existing data set. 
However, actual business results can only be measured while in operation. 

Moreover, in each business scenario some information is easy to retrieve, such as cost per token, other can be estimates, averages, or difficult to find. Moreover, different scenarios can also have different results. Each scenario can lead to the creation of a new model to explain the expected RoI and a sensitivity analysis that evaluates how changes in input parameters affect the RoI. 

In this paper, we model scenarios in which the gains and benefits are the direct result of one business operation. This is, for example, the case for a recommender system acting during an online sale, which could bring short-term gains that are very quantifiable. 
In other scenarios, listed as future work, an operation can have only an indirect impact. This would be the case for the detection of fake news in a social network that could enhance the perception of trust in its users, resulting in long-term benefits. 

The following section makes some considerations and assumptions for the models. The third section shows the model for the scenario based on the RoI of a single operation that can succeed or not. The fourth section shows the model for the scenario based on a binary classification task. The fifth section describes the model to be used when it is not possible to work with a single operation but actually with the general result of using an LLM in the business. The sixth section discusses some next steps, while the last section shows the conclusions. 

\section{Basic premises for our models}

According to the PMBoK, \nth{7} edition, ``value is the ultimate indicator of project success''. For every planned project, it is important to analyze the value of the project in different qualitative and quantitative terms. 
Among other indicators, financial results are key performance measures for project success. When establishing a business case for a project, it is crucial to understand its costs, revenues, and return on investment~\citep{PMBOK7}. 

Costs are the total amount of money spent on a project, while benefits are the total amount of money gained from a new project. Earnings are the difference between benefits and costs \footnote{We will use earnings $E$ in this paper to avoid using $P$ for profits, using $P$ for probability}, the total amount of money the project will bring to the company, and return of investment (RoI), is the relation between earnings and costs~\citep{PMBOK7}. 

If $B_p$ is the benefits of a project an $C_p$ the cost, the earnings $E$ and RoI $R$ are defined as~\citep{PMBOK7}:
\begin{align}
    E &= B_p-C_p \label{eq:earningsbasic}, \\ 
    R &= \frac{E}{C_p} \notag, \\ 
    R &= \frac{B_p-C_p}{C_p} \label{eq:roi}. 
\end{align}

Even if the project earnings are positive, it is possible that a project results in some gains $G$ and some losses $L$, therefore we can expand this model to:
\begin{align}
    E &= B_p-C_p \notag, \\ 
    B_p &= G-L \label{eq:benefits}, \\ 
    E &= G-L-C_p \label{eq:fearn}, \text{and}\\ 
    R &= \frac{G-L-C_p}{C_p} \label{eq:froi}. 
\end{align}

\subsection{How LLMs are charged}

There are different ways in which LLMs are charged. Our models assume that they are charged by token, as OpenAI does~\citep{shekhar2024optimizing}. 
However, AWS charges specific engines by the hour of the instance used~\citep{llmAWS2024}. 
Moreover, the costs of the use of proprietary machines can be calculated using traditional capital and operating expenses methods~\citep{park2013FEE}. 

If one have only the usage cost by a time interval, given the expected number of transactions per this time interval ($N_{t/i}$), the average size of a transaction $(\bar{T})$, and the cost per time interval ($C_i$), the cost per token ($C$) can be calculated as: 
\begin{equation}
C = \frac{C_i}{N_{t/i}\bar{T}} 
\end{equation}

\subsection{An anedotal model from industry}

In this subsection, we use our experience to discuss how most companies try to calculate the expected earnings from using an LLM in a project.

As our experience shows, the most commonly used method to calculate the earnings of an investment in LLM only considers the cases of success. 
The basic idea is that the LLM will be used in a large number of transactions, modeled as $N$, aiming to produce some aggregate value, or gain, modeled as $G$, only in a few of them, modeled as $M$, $M<N$. The total expenditure is $N$ multiplied by the transaction cost $C_t$

Thus, under this context, it is possible to define the earnings as:
\begin{equation}
    E = G  M  - N  C_t, 
\end{equation}

Although RoI is a textbook value, it is not common for it to be calculated. In this case it is: 
\begin{equation}
    R = \frac{G  M  - N  C_t}{N  C_t}. 
\end{equation}

We bring attention to the fact that although one is actually making a prediction, probabilistic terms such as ``expected value'' and ``probability of success'' are not commonly used in a business case for a project.

This model represents the idea that the introduction of the LLM allows the company to convert some lost revenue into new revenue; however, it disregards the fact that the result of an operation using the LLM can be bad for the business. 
Moreover, usually it does not consider business tasks, that is, it considers that if the LLM task is a success, then the business task will also be a success.

A better model  should include losses from the overall results of using LLMs, defined as $L$  and the earnings would be the following:
\begin{equation}
    E = G  M  - N  C_t - L Q, 
\end{equation}
Where $Q$, $Q\ll M$, is the number of transactions that would result in a negative result.

With the same reasoning, RoI can be recalculated as:
\begin{equation}
    R = \frac
    {G  M  - N  C_t - L  Q}  
    {N  C_t}.
\end{equation}

This model is yet too simplistic, and we improve it in the next subsections, showing how it can be made more representative of the problem.

\subsection{Choice of Costs in the Model}

We understand that using an LLM is only part of the cost of a project. 
The costs can be divided into fixed costs, $C_f$, and variable costs $C_v$~\citep{PMBOK7}. 
In internet operations, variables costs can be seen as proportional to the number of transactions $t$, therefore we use $C_v(t)$ to discuss variable costs. 

In the projects we are interested in, part of these variable costs is due to the use of an LLM, paying a fee per transaction, or using in company machines and being able to calculate the cost per transaction. 
This can be or not the dominant variable cost; however, if the project is chosen, the selection of an LLM is an important economic decision, the focus of this article. 
We will model the other variable costs as $C_o$.

Therefore, the total cost $C_T$, and variable cost $C_v(t)$ are, in our case:
\begin{align}
    C_T = C_f + C_v(t) \label{eq:totalcost}, \\
    C_v(t) = C_\text{LLM}(t) + C_o(t) \label{eq:variablecost}.
\end{align}

Moreover, since in this article we are interested in the operational costs of using LLMs, we suppose a company that is already operating its business and already decided to use LLMs in parts of its operations; however, it did not decide yet which LLM to use, due to the difference in LLM costs and performance. 
We also suppose that, for most enterprises, operational costs will be greater than development costs, which are considered fixed costs. 
Also, since the development costs are mostly similar for all LLMs, this is also a point that will not help the LLM choice.

Therefore, our premise is that we only need to analyze on $C_\text{LLM}(t)$, with all other costs considered equal for the same application, according to the principle ``ceteris paribus''. 

The cost per token value, $C$, is usually very small and is charged by thousands or millions of tokens. 
According to OpenAI, one can ``think of tokens as pieces of words, where 1,000 tokens is about 750 words''. 
OpenAI also charges tokens in the input cheaper than tokens in its output~\citep{Open-ai-pricing}. 

For example, for $1,000,000$ tokens, the OpenAI prices in May 2024 for tokens in the input were  US\$ $5.00$ for \verb|gpt-4o| and $0.50$ for \verb|gpt-3.5-turbo-0125|. 
Output tokens are usually three times more expensive. The most expensive OpenAI models charged US\$$10$ for input token and US\$$30$ for output tokens~\citep{Open-ai-pricing}. 

\subsubsection{Are there other costs?}
\label{othercosts}
A more complete analysis should use the impact of other costs, even if they are the same, to better visualize the real impacts on the business. 
The following is a consideration of other costs.

\begin{itemize}
\item Network costs: many cloud billing models include network costs, with different costs for using the service's internal network and using the external network. This may influence the choice of a service provided by the already-contracted cloud provider. Since there is a relation between transaction size and network costs, it would not be difficult to bring network costs to our model.
\item Embedding costs: If it is necessary to use RAG, or another technique that uses embedding, the cost of RAG may be important. Normally, the RAG cost is completely dominated by the most expensive models, but it is of the same order of magnitude as that of the cheapest models. For example, the embedding \verb|ada-v2| has the same cost as the model \verb|gpt-2|~\citep{Open-ai-pricing}. Embeding costs can be easily added to our model.
\item Relationship between input and output: As we have seen, as some models have different costs for input and output, it is possible that, for similar tasks, but with different outputs, the cost-benefit ratio changes. This is no challenge to our models, since it only impacts the calculation of the total cost of one transaction.
\item Refining costs: If the option is for a task where periodic refining is required, it may be interesting to consider these costs. This would bring about a more complex model including periodic costs that would actually be fixed costs in the project.
\item Reinforcement learning strategies: Along with refinement, reinforcement and learning strategies can have direct effects on cost. This would bring new variables to the model since those strategies would be classified as variable cost.
\item Discounts and ``free'' machines: it is possible that some LLMs provide discounts when transactions are bought in advance or in a large quantity, and it is also possible that available machines are underused and using them will actually lower the average price per transaction. This would bring about a more complex modeling of the cost per transaction.
\end{itemize}

In addition, there are other nonfinancial issues to take into account that affect the return, such as response time, i.e. latency, which can lead to a result, in practice, worse than the theoretical one, because a model that takes longer than expected by a user can lose the opportunity to present its results, because the user gives up navigation, generating a lower success rate, in practice, than expected.

\subsection{Why multiple scenarios}

In this paper, we choose to create different models for some different scenarios. 

Each company and even each project in a company's portfolio may have different data for its managers to analyze and predict the outcome of a project.

For example, in certain projects the gain can be calculated per operation. For example, a recommendation system that recommends a single item for purchase in the last step of a sale can be built based on the premise that it will increase the ticket in US\$ 10 in 10\% of the cases, but it will also lose the complete sale transaction, due to missing the user's focus on the sale, in 1\% of the cases. 
A system that avoids toxicity in a game can be created on the premise that it will reduce churn by 5 percentage points. 
Although the first example deals with the problem based on the cost of an operation, the second can only be seen from a global scenario of the company, as each transaction will not directly influence the churn rate, but rather the global change in the spirit of in-game conversations that indirectly attracts or repels clients.

\section{A Decision Theoretic Model}

During the planning of a project, the earnings and RoI can only be estimated, as there are uncertainties and risks in projects~\citep{PMBOK7}.
Therefore, any decision should take these estimates into account. Therefore, we now model the expected value of earnings and RoI, $\ex{E}$ and $\ex{R}$~\citep{frahm2019RCSC}.

Moreover, to allow for a better decision in theoretical and practical perspectives, one needs to establish a model that includes costs, gains, and losses from the use of an LLM and also a probabilistic view of success. 
In a decision-theoretic approach, this is possible, among other ways, by using Savage's Theory of Rational Choice~\citep{Savage1972FS}, as described by~\citet{frahm2019RCSC}.

According Savage's theory, the expected value for the utility of an state $s$ is:
\begin{equation}
    \ex{u(s)} \equiv \ex{f}\coloneqq \sum^n_{i=1}{p_iu_i},
\end{equation}
where $p_i$ is the probability that an action (or function) $f$ will lead to a state of utility $u_i$. 

We start by considering the action of using an LLM to solve a business problem or improve a business situation, with only two possible outcomes, or states: success of failure of a business transaction. 
We initially model the probability of success as $P$ and the probability of failure as $(1-P)$. Therefore the expected value (in earnings) and expected RoI of a project:
\begin{align}
    \ex{E} 
 &= GP-L(1-P)-C_p \label{eq:probearn}, \\
   \ex{R} 
 &= \frac{GP-L(1-P)-C_p}{C_p}. \label{eq:probroi} 
\end{align}

These decision-theoretic equations~\ref{eq:probearn} and \ref{eq:probroi} will guide this article. For each proposed model, we will discuss how the variables $G$, $L$, $P$ and $C_p$, in the form of cost per token $C$, affect $E$ and $R$.

\subsection{How to model success}

In this work, we start by using the probability of success of an LLM in a business task as $P$. However, the analysis of such probability is not simple, due to the fact that there are actually two probabilities involved: the probability of an LLM to fulfill its task
and the probability of the task having an effect on the business. 
For example, even if an LLM detects that users are prone to include a product in their basket, and a sales system on the Internet provides this suggestion to the users, they can decide not to or they be distracted by the offer and do not close the operation, resulting in a loss.

In this paper, we consider both probabilities as aggregated probabilities. 
However, we draw attention to the fact that the second probability is conditional to the first. 
Therefore, being $P_b$ the probability of business success and $P_t$ the probability of the specific LLM task success, one could have the basic assumption that:
\begin{equation}
    P = P(P_b | P_t)(P_t).
\end{equation}

Moreover, a more realistic model should include the chance of business success when the LLM task fails. For example, it is possible that the LLM suggests a wrong product and, by chance, the customer accepts it. This can be modeled as:
\begin{equation}
   P = P(P_b | P_t)P_t + P(P_b | (1-P_t))(1-P_t). \label{eq:realP}
\end{equation}

Finally, one can also model the probability of a correct prediction from the LLM to lead the user to a path where he does not complete the transaction. In this case, one should also model the probability of losing the transaction. We do that in \autoref{sec:TPTN}. 

The values of $P$, $P_b$, and $P_t$ can be estimated in different ways and will be discussed elsewhere. For example, $P_t$ can be estimated by creating a tagged dataset, $P$ can be estimated, even without knowledge of $P_t$ and $P_b$, with tests in operation, such as A/B tests, and $P_b$ can be estimated from similar operations in other business channels made by humans or by calculating indirectly from $P$ and $P_t$.

\section{A Model for Commercial Operations Based on a Single Transaction}

This scenario supposes that a company evaluates different LLMs for the same task, based on the probability of success of each LLM in the business. 
The analytical steps include building a decision-theoretic RoI model, performing a qualitative sensitivity analysis based on first-degree partial derivatives, and performing a Sokol evaluation of the first and second degree. 
In this scenario, we also show a numerical example based on real prices from the market in May 2024, and an arbitrarily supposed performance that is compatible with LLMs with the specified cost, to show how RoI formulae can be used in a real case.

Suppose that a company evaluates an LLM, such as \verb|gpt-4|, for a specific business task.
The costs associated with the use of the LLM are expressed in $C_{LLM}$ dollars per $K$ tokens. Therefore, the cost of a token is $C=\frac{C_{LLM}}{K}$
The cost of the transaction can be modeled as $C_t =CT$
where  $C$  is the cost of a token and $T$ is the average size of the transactions in tokens. 

In this first analysis, we consider only the (final) probability of success, modeled as $P$. 

To quantify the economic impact of choosing an LLM, it is assumed that success in the task generates a gain \(G\) and an error implies a cost \(L\). 
Gains and losses are inherent to the task and not to the LLM. 

Thus, the expected value of earnings, $\ex{E}$, and RoI, $\ex{R}$, per transaction using an LLM, can be calculated with a decision-theoretical model where the earnings are calculated from the gain per transaction that generates a positive result multiplied by the probability of the result being positive, minus the loss per transaction that generates a negative impact, multiplied by the probability of a negative impact occurring and minus the total cost of the transaction, as in \autoref{eq:earnings}. The expected RoI is the expected earnings divided by the total cost of the transaction, as in \autoref{eq:operroi}.

\begin{align}
\ex{E} &= \left(G - C T\right) \cdot P - \left(L + C T\right) \cdot \left(1 - {P}\right) \nonumber \\
\ex{E} &= GP - L(1 - P) - CT \label{eq:earnings}
\end{align}

\begin{align}
\ex{R} &= \frac{(G - CT) \cdot P - (L + C T) \cdot (1 - P)}{C T}  \nonumber \\ 
\ex{R} &= \frac{GP-L(1-P)}{CT}-1 
\label{eq:operroi}
\end{align}

These expressions facilitate a comparative analysis of earnings and RoI, taking into account not only the direct costs of using LLMs but also the economic consequences of the results produced by these technologies.

\subsection{Example of Using the Model}

Suppose that the candidate LLMs are \verb|llm-1| and \verb|llm-2|, whose values in May 2024 for input were $C_{\text{llm-1}} = 10$ dollars per million tokens, $C_{\text{llm-2}} = 0.5$ dollars per million tokens. 
The example task uses $T=1000$ tokens in the input and can return an average gain of US\$ 10, i.e.,\(G = 10\), or an average loss of US\$ 1, i.e.,\(L = 1\) dollar. Finally, suppose that, for this specific task, the tests show that \(P_{\text{llm-1}} = 95\%\) and \( P_{\text{llm-2}} = 80\%\).

For classification tasks, the cost can be completely dominated by the input, as the output can only be 1 token. Under the assumption of those specific conditions in a classification task, the earnings ($E$) and the  RoI ($R$) for each LLM would be:

\begin{align*}
\ex{E}_{\text{llm-1}_i} &= (10 - \frac{10 \cdot 1,000}{1,000,000}) \cdot 0.95 + (-1 - \frac{10 \cdot 1,000}{1000000}) \cdot ( 1 - 0.95) \\
\ex{E}_{\text{llm-1}_i} &= 9.44 \text{ dollars} \\
\ex{R}_{\text{llm-1}_i}&= 944 \\
\ex{E}_{\text{llm-2}_i} &= (10 - \frac{0.5 \cdot 1,000}{1,000,000}) \cdot 0.80 + (-1 - \frac{0.5 \cdot 1,000}{1,000,000}) \cdot (1 - 0.80) \\
\ex{E}_{\text{llm-2}_i} &= 7.79950 \text{ dollars}\\
\ex{R}_{\text{llm-2}_i} &= 15,599 
\end{align*}

This example demonstrates that, despite \verb|llm-1| costing 20 times more and presenting a higher income, it also presents a significantly lower return on investment. This is a typical result of comparing projects with a large difference in investment~\citep{PMBOK7}.  

However, if we consider a very large transaction, we can change the value of $T$ to $128,000$ and maintain the other values.
With these data, it is possible to find an approximate value of $84\%$ as the minimum performance required for \verb|llm-2| to become more economical than \verb|llm-1|.

\begin{figure}[hbt]
\centering
\includegraphics[width=0.7\textwidth]{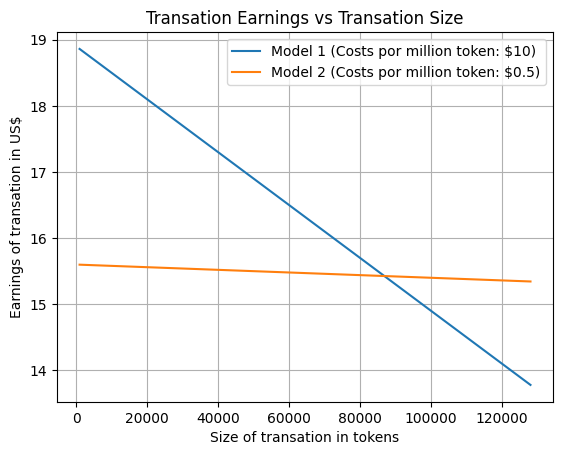}
    \caption{Transaction gain variation as a function of size for two models with different performance (95\% and 80\%) and cost (US\$10 and US\$1).}
    \label{fig:evolve}
\end{figure}

The graph in \autoref{fig:evolve} shows the evolution of the earnings for two competitive models according to the values used in the previous example as a function of the average transaction size.

\subsection{Model Analysis}

We must point out that due to the use of LLM in a commercial context, the uncertainties will be mainly in $G$, $L$, and $P$. It would be reasonable to expect a normal probability distribution for $P$ and a normal or even power probability distribution for $G$ and $L$. Since $G$ and $L$ are business values, and $P$ is the business result that comes from performing the AI tasks correctly, these values are not under the direct control of the project team. We recall that a complete model of the probability of success should start from \autoref{eq:realP}, but this is beyond the scope of this article.

On the other hand, $T$, that is, the size of the transaction in tokens, it is more under the control of the project team, since it depends on the size of data used, and desired, at input and output, and also the size of the prompt.
One should also expect a normal or power probability distribution for $T$. Finally, for a single LLM chosen, $C$ is actually a constant, and we only need to understand its impact on the decision when more than one LLM is considered. 

Looking at \autoref{eq:earnings}, we can also see that if $G$ and $L$ are large and $T$ small, the factor $CT$ will be small and $\ex{E}$ will be mainly a function of $GP-L(1-P)$, and since $G$ and $P$ are expected to have the same distribution for a single business task, $P$ would have the greater impact. This could greatly impact the evaluation of prompt compression techniques, such as LLMLingua\citet{jiangetal2023llmlingua,jiangetal2023long,pan2024llmlingua2} for small prompts. 
For example, suppose that a classification task using a $1,000$ token transaction could result in US\$$10$ gain or US\$$1$ loss. Its cost would be US\$$0.005$ in \verb|GPT 4o|. Therefore, compressing the prompt $20\times$ would save US\$$0.00475$ in $CT$. Meanwhile, each $1\%$ lost in $P$ would result in a $GP-L(1-P)$ loss of US\$$0.11$.  At the other hand, for small $G$ and $L$, and larger $T$, compression could be effective.

Taking all this into account, we decided to perform both a local and a global sensitivity analysis to understand the impact of the parameters in $\ex{E}$ and $\ex{R}$. 
We understand that the local sensitivity is easier to understand and reasonable in this case, while the global sensitivity is theoretically stronger; therefore, we also analyze the linearity of the functions to understand how valid the local analysis is.

\subsubsection{Local Sensitivity Analysis}

This first sensitivity analysis is local and qualitative. It is carried out by evaluating the partial derivatives of \( \ex{E} \) in relation to each variable in their respective intervals. This provides insight into how sensitive \( \ex{E} \) is to changes in each parameter within the ranges of variation.

Taking into account \autoref{eq:earnings}, it is possible to find all the partial derivatives for variables $G$, $C$, $T$,  $P$ and $L$. 

The partial derivatives for $\ex{E}$ are:
\begin{align}
\frac{\partial \ex{E}}{\partial G} &= P, \label{eq:partialg} \\
\frac{\partial \ex{E}}{\partial L} &= P-1, \label{eq:partiall} \\
\frac{\partial \ex{E}}{\partial C} &= -T, \label{eq:partialc} \\
\frac{\partial \ex{E}}{\partial T} &= -C, \label{eq:partialt} \\
\frac{\partial \ex{E}}{\partial P} &= G + L.\label{eq:partialp}
\end{align}
The Hessian of $\ex{E}$ is mostly composed of zeros, indicating a fairly linear behavior that would be very acceptable for a local analysis. 

Similarly, the partial derivatives for $\ex{R}$ are:
\begin{align}
\frac{\partial \ex{R}}{\partial G} &= \frac{P}{CT}, \\
\frac{\partial \ex{R}}{\partial P} &= \frac{G + L}{CT}, \\
\frac{\partial \ex{R}}{\partial L} &= -\frac{(1-P)}{CT}, \\
\frac{\partial \ex{R}}{\partial C} &= -\frac{(GP - L(1-P))}{C^2T}, \\
\frac{\partial \ex{R}}{\partial T} &= -\frac{(GP - L(1-P))}{CT^2}.
\end{align}
Thus, the Hessian matrix \(H\) of \(\mathbb{E}[R]\) with respect to variables \(G\), \(C\), \(T\), \(L\), and \(P\) is:
\[
H = \begin{pmatrix}
0 & -\frac{P}{C^2 T} & -\frac{P}{CT^2} & 0 & \frac{1}{CT} \\
-\frac{P}{C^2 T} & 2 \frac{GP - L(1 - P)}{C^3 T} & \frac{GP - L(1 - P)}{C^2 T^2} & 0 & -\frac{G + L}{C^2 T} \\
-\frac{P}{CT^2} & \frac{GP - L(1 - P)}{C^2 T^2} & 2 \frac{GP - L(1 - P)}{C T^3} & 0 & -\frac{G + L}{CT^2} \\
0 & 0 & 0 & 0 & -\frac{1}{CT} \\
\frac{1}{CT} & -\frac{G + L}{C^2 T} & -\frac{G + L}{CT^2} & -\frac{1}{CT} & 0
\end{pmatrix}.
\]
The Hessian matrix of $\ex{R}$ has fewer zeros, and the small value of $C$, found in the denominator of most partial derivatives, will bring greater values. This reflection is a challenge for local sensitivity analysis; however, looking at the partial derivatives still allows for some advantages on interpretability.

From these results, we can see that there is a strong interaction between $G$, $P$ and $L$, and also between $T$ and $C$ in earning. Regarding RoI, however, we see the importance of $CT$ as a product. We also call attention to the fact that since $C$ is usually very small, in the order of $5\times10^{-6}$ or event $10^{-7}$, for small values of $T$, and large values of $G$ and $L$, the value of $CT$ can be of no importance in $\ex{E}$. 

In addition, it is clear that the sensitivity of both $\ex{E}$ and $\ex{R}$ to $P$ is influenced by $G+L$. This shows the importance of considering not only gains but also losses.

The terms $CT$, $C^2T$ and $CT^2$ that appear in the partial derivatives for RoI show how the inversely affect the importance of all other variables.

It is reasonable to consider the following ranges for parameters: \(1 \leq G \leq 1000\), \(0 \leq L \leq 1,000\), \(0.01 \leq C \leq 100\), \(0.75 \leq P \leq 1\), and \(10 \leq T \leq 128,000\).

\subsubsection{Global Sensitivity analysis of earnings by Sobol technique}

Although local sensitivity analysis looks at one variable at each time in the neighborhood of a point, and thus is only adequate, if so, for linear and mostly linear functions, global analysis can study the effect of all variables at the same time, for specific distributions, even in the case of nonlinearity~\citep{saltelli2004sensitivity,saltelli2008global} . 

\citet{sobol2001global} proposed a method for global sensitivity analysis based on running a model with a sampling generated under certain premises, and make a variance decomposition that allow the calculation of a set of indexes of sensitivity. 

The first-order Sobol index measures the individual contribution of each input variable to the variance in the model output, ignoring interactions with other variables. In simpler terms, it quantifies how much of the uncertainty in the model output can be directly attributed to changes in a specific input, while all other inputs are held fixed. This quantification is performed between values $0$ and $1$~\citep{sobol2001global}. 

The total order Sobol index quantifies the overall contribution of an input variable to the output variance, considering both its individual effect and its interactions with other variables. Essentially, it measures the impact of a variable on the output by accounting for all possible ways in which it can influence the result, including its combined effects with other variables. This index also ranges between values \(0\) and \(1\), indicating the proportion of output variability attributed to the total influence of the given input variable~\citep{sobol2001global}.

Finally, the second-order, and higher orders, Sobol index is an extension of the first-order index and is used to measure the combined effect or interaction between pairs, or tuples, of input variables on the variance of the model output. While the first-order index focuses on the individual contribution of each variable, the second-order index examines how the interaction between two variables contributes to uncertainty in the output, beyond what would be expected based on the individual effects of each variable~\citep{sobol2001global}. This allows for a good visual representation of the combined effect of pair of variables in a matrix.

Therefore, a Sobol sensitivity analysis~\citep{sobol2001global} was performed, using the SALib package~\citep{Iwanaga2022,Herman2017}, in Python, using the ranges for the \autoref{eq:earnings} variables described in \autoref{tab:variablespancom}. We used $2^{20}$ samples in each execution. 

\begin{table}[hbt]
\centering
\begin{tabular}{cccc}
\toprule
\textbf{Variable} & \textbf{Minimum} & \textbf{Maximum} \\
\midrule
G & $1$ & $1,000$ \\
L & $0$ & $1,000$ \\
C & $0.01$ & $100$ \\
P & $0.10$ & $1$ \\
T & $50$ & $128,000$ \\
\bottomrule
\end{tabular}
\caption{Limits used for each variable in Sobol analysis for the commercial operation.}
    \label{tab:variablespancom}
\end{table}

\begin{figure}[!htb]
\begin{subfigure}{.48\textwidth}
\centering
\includegraphics[width=0.95\textwidth]{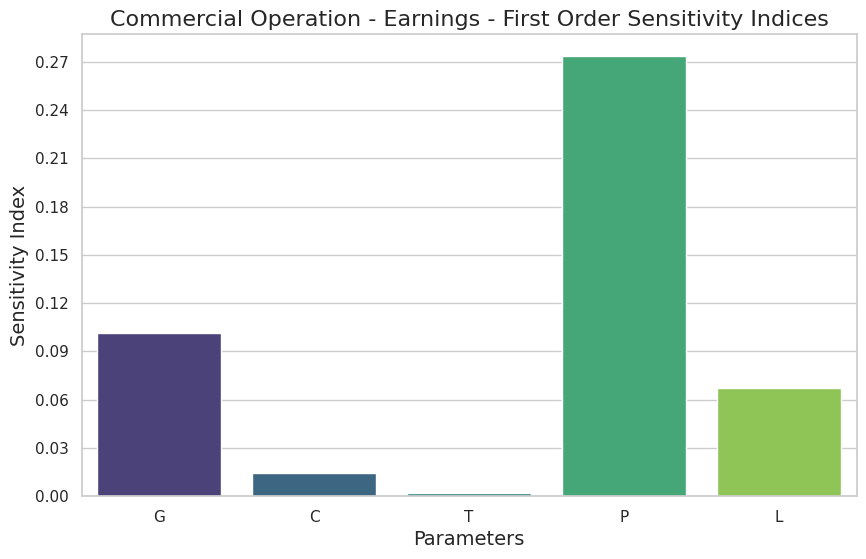}
\caption{First-order Sobol index}
\label{fig:co_e_fo}
\end{subfigure}%
\hfill
\begin{subfigure}{.48\textwidth}
\centering
\includegraphics[width=0.95\textwidth]{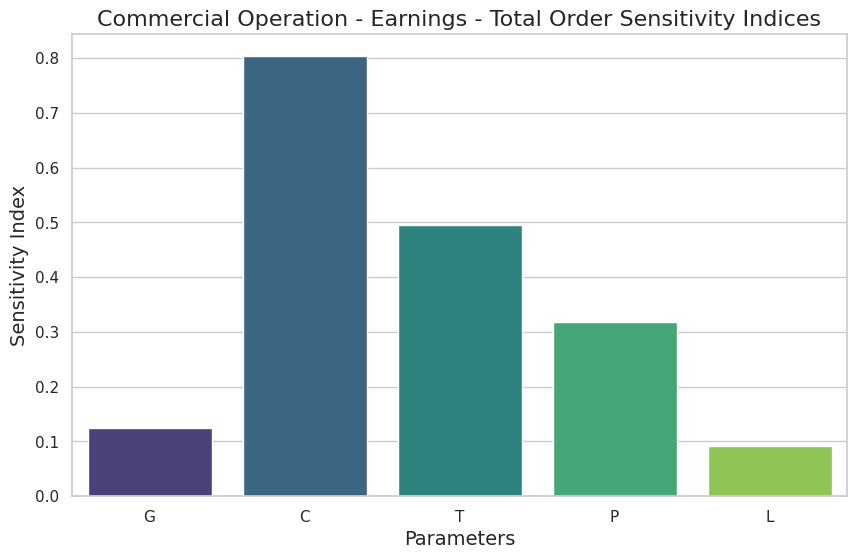}
\caption{Total-order Sobol index}
\label{fig:co_e_to}
\end{subfigure}\\[1ex]
\caption{Global sensitivity analysis of earnings in a commercial operation.}
\label{fig:co_e}
\end{figure}

\begin{figure}
\centering
\includegraphics[width=0.7\textwidth]{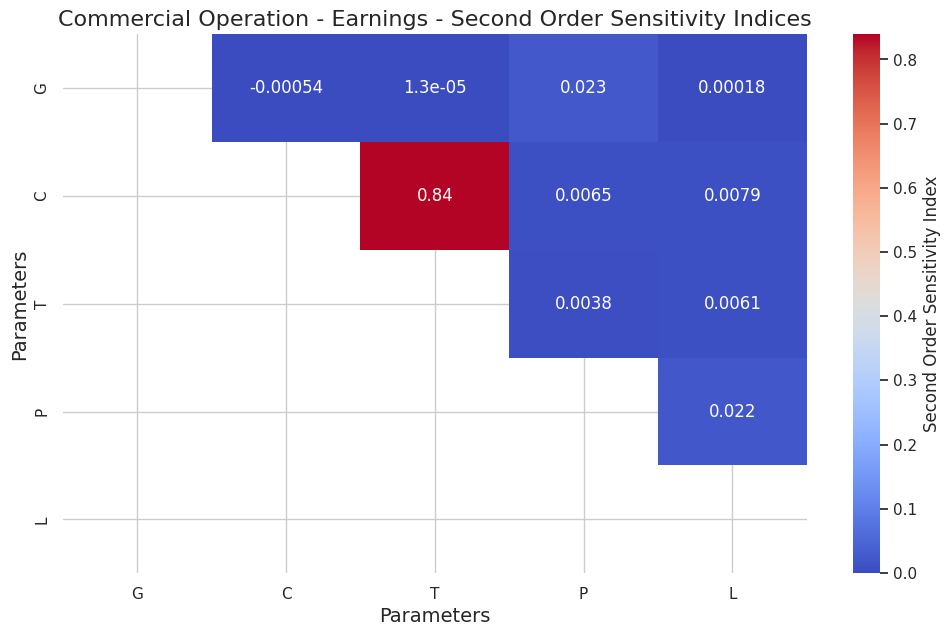}
    \caption{Global sensitivity analysis using the second-order Sobol index of earnings in commercial operations.}
\label{fig:co_e_so}
\end{figure}

From the global sensitivity analysis of the earnings for the commercial operation model, it can be seen that the value of $P$ is clearly the most important factor in the first order analysis \autoref{fig:co_e_fo}. For the total order index, \autoref{fig:co_e_to} shows the importance of $C$ and $T$, which is also reflected in the second order values, shown in \autoref{fig:co_e_so}, which shows the clear predominance of the pair $C$ and $T$ as a pair. The product $CT$ is the total cost of the transaction.

\begin{figure}[!htb]
\begin{subfigure}{.48\textwidth}
\centering
\includegraphics[width=0.95\textwidth]{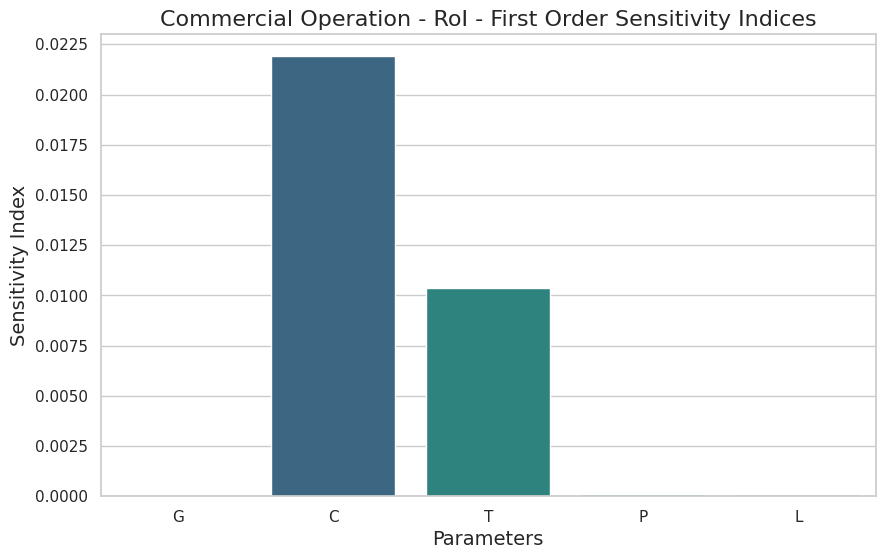}
\caption{First-order Sobol index}
\label{fig:co_r_fo}
\end{subfigure}%
\hfill
\begin{subfigure}{.48\textwidth}
\centering
\includegraphics[width=0.95\textwidth]{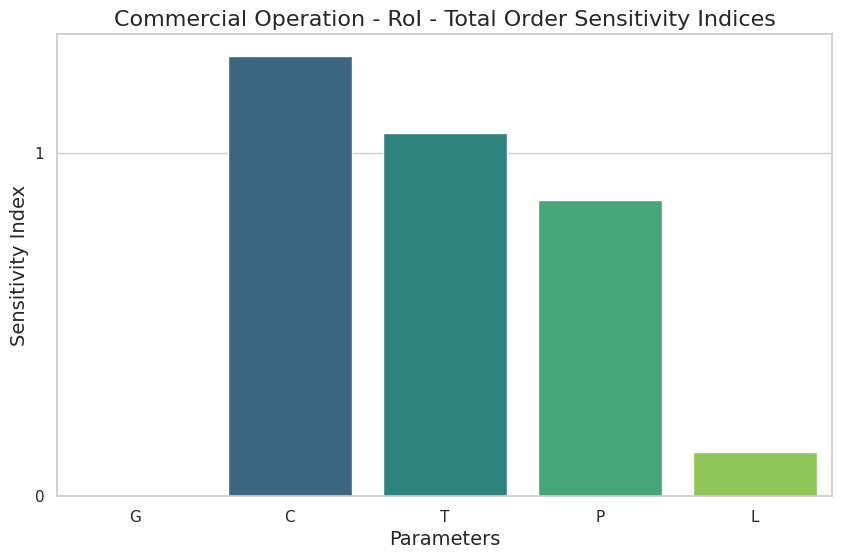}
\caption{Total-order Sobol index}
\label{fig:co_r_to}
\end{subfigure}\\[1ex]
\caption{Global sensitivity analysis of RoI for commercial operations.}
\label{fig:co_r}
\end{figure}

\begin{figure}
\centering
\includegraphics[width=0.7\textwidth]{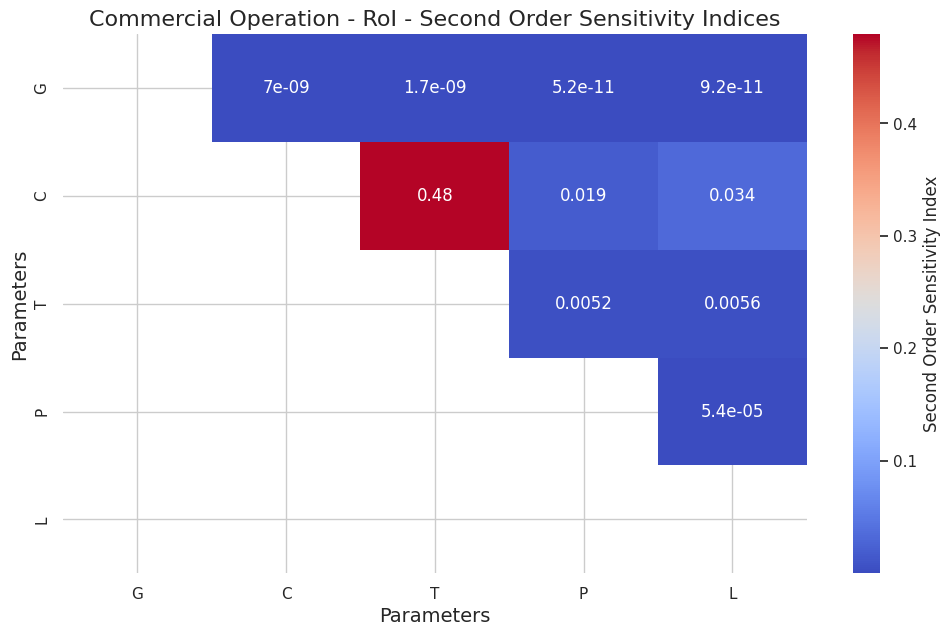}
    \caption{Global sensitivity analysis of RoI using the second-order Sobol index for commercial operations}
\label{fig:co_r_so}
\end{figure}

The RoI sensitivity analysis for the commercial operation model shows that RoI is particularly sensitive to changes $C$, $T$ and $P$ for the total order, within the given ranges. In particular, the combination of $C$ and $T$ has the greater impact on the variation of RoI of the second order. Since $C$ and $T$ were already the most important factors in the earnings analysis, it is expected to see their importance in RoI, where they play a greater role in the final value of the RoI equation.

Earnings are basically sensitive to $C$ and $T$, with $P$ playing an important role.
Based on this scenario, one should pay attention to a good prediction of $T$ and $P$, since $C$ would be fixed for a specific LLM.

\subsection{Discussion of the RoI vs. Earning dillema}

Selecting between projects that differ in their potential earnings and RoI is a common dilemma faced by project managers~\citep{PMBOK7}. The choice between a project with smaller earnings but higher RoI and another with larger earnings but lower RoI involves a strategic decision-making process that considers various financial and non-financial factors. 

Although RoI is a crucial metric, the absolute earnings potential of a project should not be overlooked. A project with lower RoI but higher total earnings might be more beneficial if the additional revenue significantly impacts the company's financial health or growth prospects. However, under budget constraints, projects with lower costs and higher RoI might be more feasible, even if their total earnings are lower. Actually, since it is usually easy to swap LLMs, this can be a bootstrap approach. Analysis of current and future cash flow could help with the decision. 

\section{Modelling of a Binary Classification Problem}
\label{sec:TPTN}
A binary classification problem is a type of supervised learning task in machine learning where the goal is to predict to which of two classes (categories) a particular instance belongs to~\citep{bishop2006}. 
The output variable in binary classification is typically a categorical variable with two possible values, often represented as 0 and 1, ``yes'' and ``no'', or ``true'' and ``false''.

For example, in a medical diagnosis application, the task might be to classify whether a patient has a particular disease or not based on various input characteristics such as age, blood pressure, cholesterol levels, and other health indicators. 

In the context of binary classification, it is crucial to understand the different outcomes that can occur. Usually, the results of binary classification problems are described through a confusion matrix, which lists true positives, false positives, true negatives, and false negatives~\citep{bishop2006}. 

A true positive (TP) occurs when the model correctly predicts the positive class. For example, if the model identifies a transaction as fraudulent and is indeed fraudulent, this is a true positive. In contrast, a true negative (TN) occurs when the model accurately predicts the negative class, such as correctly identifying a non-fraudulent transaction~\citep{bishop2006}.

A false positive (FP) occurs when the model incorrectly predicts the positive class. In this case, the model flags a non-fraudulent transaction as fraudulent, leading to unnecessary checks and associated costs, represented by $\lfp$. Finally, a false negative (FN) occurs when the model fails to identify a positive instance, such as missing a fraudulent transaction. This oversight can result in significant losses, represented by $\lfn$. Each of these outcomes has a specific impact on the cost and benefit analysis of the classification system~\citep{bishop2006}.

The costs involved with the binary classification operation require a more detailed calculation. 
First, it must be delimited to what will be done with each operation. 
Furthermore, considering only the classification as true or false, one must consider the costs related to all possible options between what to do true positives, true negatives, false positives and false negatives.

At first, it is interesting to analyze a change in our representation to consider the different costs. 

With this modifications, the expected earnings can be modeled as:
\begin{equation}
\ex{E}=(G-CT)\pvp -CT\pvn -(\lfn +CT)\pfn -(\lfp +CT)\pfp \label{eq:earnbc},
\end{equation}
where $\pvp $ is the probability of being a true positive, which brings the gain $G$, $\pvn $ is the probability of being a true negative, which brings neither gain nor loss, but has the transaction cost, $\pfn $ is the probability of a false negative, which has an additional cost $_{FN}$ (which in some cases can be considered as the loss of value $G$), and $\pfp$ is the probability of a false positive, which implies a loss $\lfp$.

From equation \autoref{eq:earnbc} we can calculate RoI as:
\begin{align}
    \ex{R}=\frac{GP_{vp}-L_nP_{fn}-L_pP_{fp}}{CT}-1\label{eq:roibc}.
\end{align}

For example, suppose that true was detected and the real class was actually true, i.e., a true positive. This can lead to a gain $G$. If a true instance occurred without being detected, a false negative indicates a loss $\lfn$. However, a false instance detected as true, a false positive, also generates another loss, $\lfp$. Finally, true negatives do not influence the business, as there is nothing to do and there is only the cost of the transaction.

\subsection{Local Sensitivity Analysis}

A direct analysis on partial derivatives is not absolutely correct, because there is the restriction that:
\begin{equation}
\pvp +\pvn +\pfp +\pfn = 1.\label{eq:sumone}
\end{equation}

The solution here is to substitute for one of these variables, and we select $\pvn$ because that is the variable that we consider to bring no gain or loss. Therefore, we apply the substitution $\pvn =1-(\pvp +\pfp +\pfn )$. Therefore, the equations 
results in:
\begin{align}
\ex{E}=(G-2CT)\pvp -(\lfn+2CT)\pfn -(\lfp+2CT)\pfp - CT, \\
\ex{R}=\frac{(G-2CT)\pvp -(\lfn+2CT)\pfn -(\lfp+2CT)\pfp - CT}{CT}.
\end{align}

This leads to partial derivatives:
\begin{align}
\frac{\partial \ex{E}}{\partial G} &= \pvp, \\ 
\frac{\partial \ex{E}}{\partial \lfn } &= -\pfn, \\
\frac{\partial \ex{E}}{\partial \lfp } &= -\pfp, \\
\frac{\partial \ex{E}}{\partial \pvp } &= G-2CT, \\
\frac{\partial \ex{E}}{\partial \pfn } &= -\lfn-2CT, \\
\frac{\partial \ex{E}}{\partial \pfp } &= -\lfp-2CT, \\
\frac{\partial \ex{E}}{\partial C} &= -T(2\pvp+2\pfp+2\pfn+1)\label{eq:cebin},\\ 
\frac{\partial \ex{E}}{\partial T} &= 
-C(2\pvp+2\pfp+2\pfn+1).
\label{eq:tebin}
\end{align}
Again we checked the Hessian matrix to analyze the linearity and found 48 zeros for 64 partial derivatives $\ex{E}$, which is fairly linear. However, again the Hessiam matrix for RoI does not show a large number of zeros, so we will skip the local analysis for RoI.

There is a point to note in equations \ref{eq:cebin} and \ref{eq:tebin}, since they can be rewriten as:
\begin{align}
\frac{\partial \ex{E}}{\partial C} &= -T(2(1-\pvn)+1), \nonumber \\ 
\frac{\partial \ex{E}}{\partial C} &= T
(2\pvn-3), \\
\frac{\partial \ex{E}}{\partial T} &= C
(2\pvn-3).    
\end{align}
The two previous equations give an interesting insight: $\pvn$ positively impacts the sensitivity to $C$
and $T$. 

From those previous equations, we see now the separation of influences: while when we used only $P$ and $(1-P)$ as measures of success and failures, there was a strong interaction between $P$ and $G+L$. Now there is a separation, $G$ and the different $L$ are mostly influenced by their own probabilities of happening, although there is always the \autoref{eq:sumone} to show interactions among the probabilities. 

\subsection{Global Sensitivity Analysis by Sobol Method}

A new sensitivity analysis can be performed. However, since all four probabilities must sum one, we made some decisions. First, we will not analyze the sensitivity to $\pvn$. Second, to avoid the difficulties of a general probalistic model, we will suppose a case where the true positives are at most 30\% and the false results are both at most 10\%. This is a sensible model because the positive cases, reflected in the true positive and false negatives, cannot be imagined to be a dominant case because, if they were, the case for using AI would be much weaker, since the challenge of finding a positive case would be small.

Using the variation of variables in the values of \autoref{tab:soboltox}, it is possible to perform the Sobol analysis again.

\begin{table}[hbt]
\centering
\begin{tabular}{cccc}
\toprule
\textbf{Variable} & \textbf{Minimum} & \textbf{Maximum} \\
\midrule
$G$& 1& 1000 \\
$\lfp $& 0& 1000 \\
$\lfn $& 0& 1000 \\
$C$& 0.01 & 100 \\
$\pfp $& 0 & .1 \\
$\pfn $& 0 & .1 \\
$\pvp $& 0 & .3 \\
T & $50$ & $128,000$ \\
\bottomrule
\end{tabular}
    \caption{Limits used for each variable in the Sobol analysis for the binary classification problem.}
\label{tab:soboltox}
\end{table}

\begin{figure}[!htb]
\begin{subfigure}{.48\textwidth}
\centering
\includegraphics[width=0.95\textwidth]{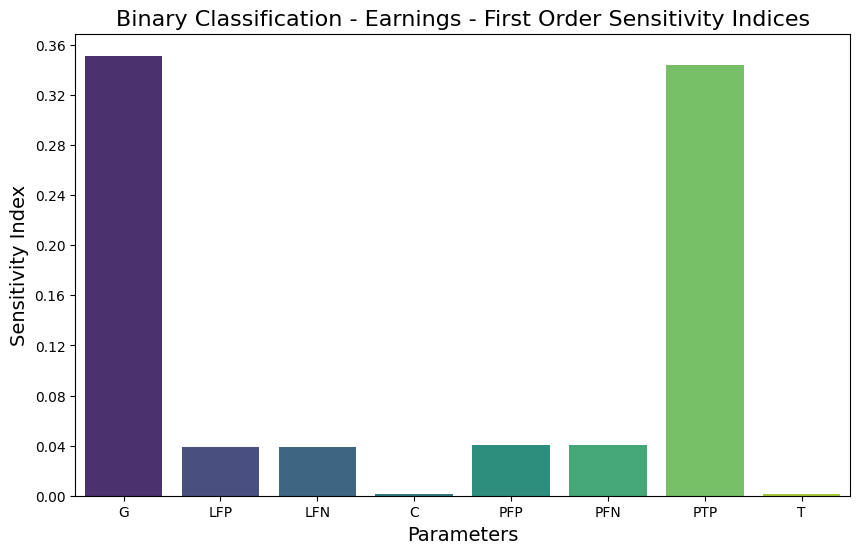}
\caption{First-order Sobol analysis}
\label{fig:bc_e_fo}
\end{subfigure}%
\hfill
\begin{subfigure}{.48\textwidth}
\centering
\includegraphics[width=0.95\textwidth]{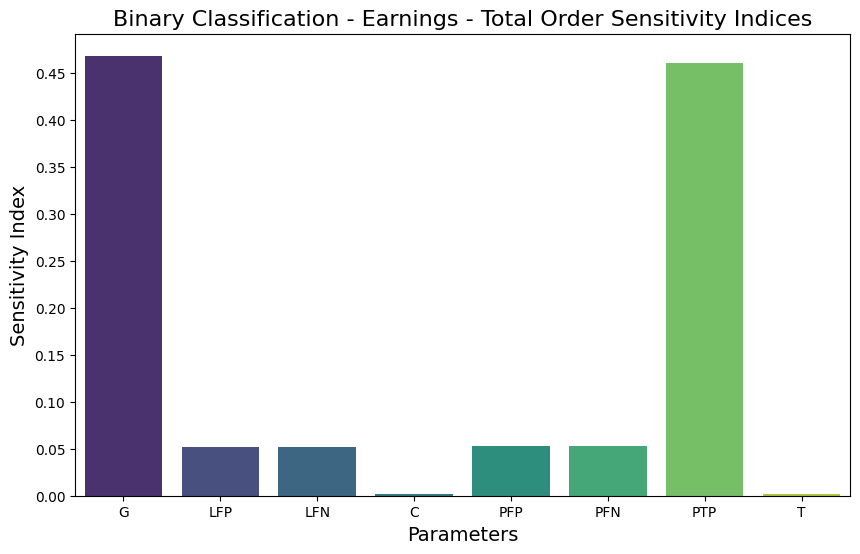}
\caption{Total-order Sobol analysis}
\label{fig:bc_e_to}
\end{subfigure}\\[1ex]
\caption{Global sensitivity analysis of earnings for the binary classification problem.}
\label{fig:bc_e}
\end{figure}

\begin{figure}
\centering
\includegraphics[width=0.8\textwidth]{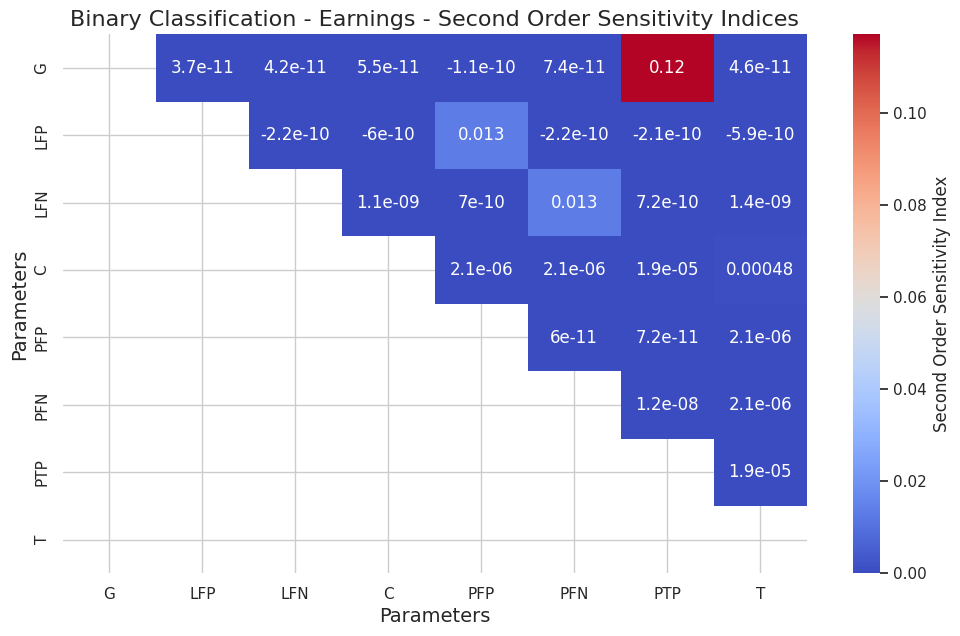}
\caption{Second order Sobol analysis for earnings for the binary classification problem.}
\label{fig:bc_e_so}
\end{figure}

\begin{figure}[!htb]
\begin{subfigure}{.48\textwidth}
\centering
\includegraphics[width=0.95\textwidth]{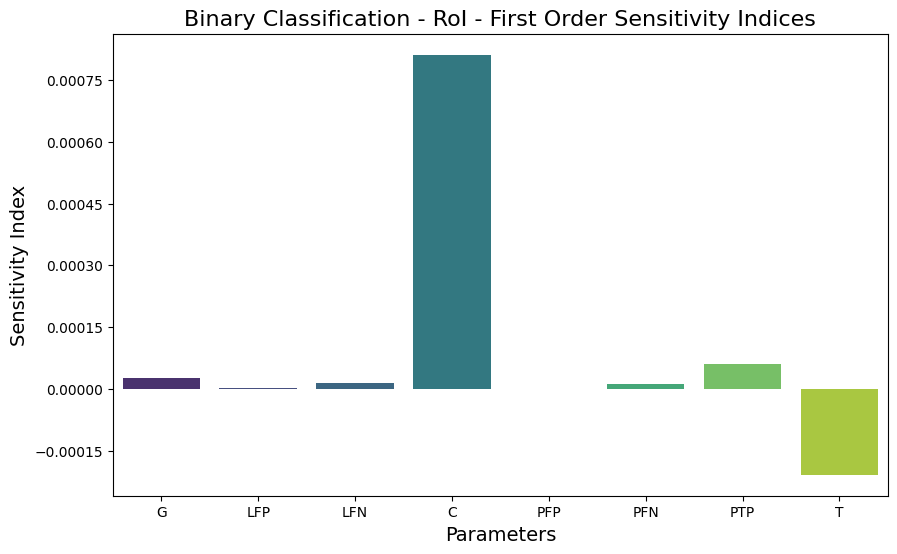}
\caption{First-order Sobol analysis}
\label{fig:bc_r_fo}
\end{subfigure}%
\hfill
\begin{subfigure}{.48\textwidth}
\centering
\includegraphics[width=0.95\textwidth]{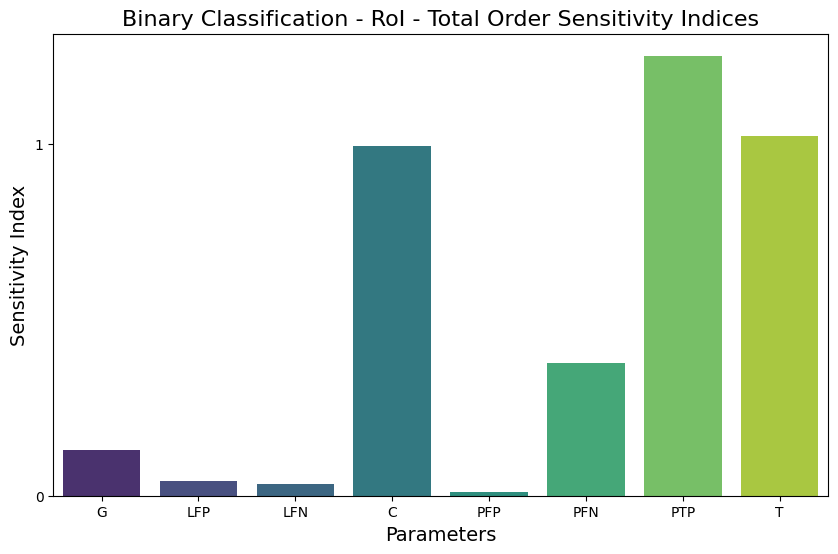}
\caption{Total-order Sobol analysis}
\label{fig:bc_r_to}
\end{subfigure}\\[1ex]
\caption{Global sensitivity analysis of RoI for the binary classification problem.}
\label{fig:bc_r}
\end{figure}

\begin{figure}
\centering
\includegraphics[width=0.8\textwidth]{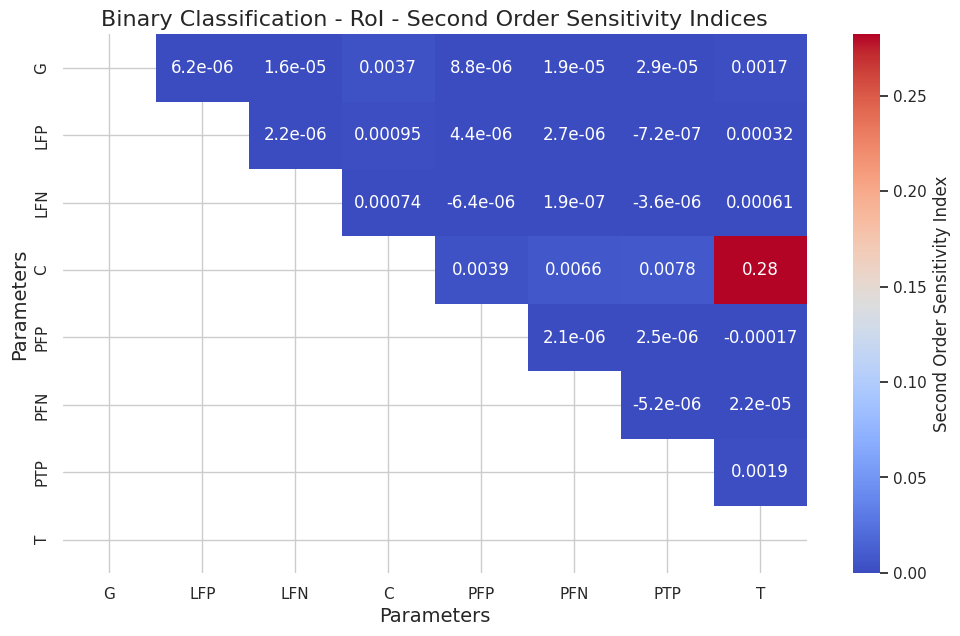}
\caption{Second order Sobol analysis of RoI for the binary classification problem.}
\label{fig:bc_r_so}
\end{figure}

From the global sensitivity analysis of the earnings in the binary classification problem, we can see the importance of $\pvp$ and $G$ in the three indexes, represented in Figures \ref{fig:bc_e_fo}, \ref{fig:bc_e_to} and \ref{fig:bc_e_so}, which clearly agree in their results. This shows the importance of a good algorithm in this case. However, costs have a small influence on earnings. 

Finally, looking at the RoI indexes, in Figures \ref{fig:bc_r_fo}, \ref{fig:bc_r_to} and \ref{fig:bc_r_so}, we see that $C$ and $T$ assume prominent roles in all indexes, while $\pvp$ has the highest total order index. Again, since the role of $C$ and $T$ is greater in RoI, this calls attention to the importance of the total cost of transaction.

\section{Related Work}

As we stated earlier, most of the work on LLM is based on their performance under specific benchmarks~\citep{chang2023survey,zhao2023survey}, disregarding the financial aspects relevant to a business operation.  

Our work aligns with the broader discourse on the return on investment in AI technologies. Recently, two articles took a similar path as ours, but with different approaches.

\citet{Gupta2024}, a conceptual approach to the RoI disccusion, propose a structured framework called the AI Maturity Continuum, which delineates the uniqueness of the RoI curve in AI projects. This continuum is segmented into three phases: problem understanding and data preparation, model improvement, and maximizing system capabilities. The first phase emphasizes understanding the problem, defining success metrics, and preparing the dataset, which parallels the initial steps in our decision-theoretic model where we define the problem and consider the costs and benefits of using different LLMs. The second phase focuses on the transition from a basic model to an optimized model, reflecting our approach of evaluating LLMs based on their performance and associated costs. Finally, the third phase involves pushing the AI system's capabilities to their limits, similar to our sensitivity analysis and fine-tuning of cost variables to maximize RoI. \citet{Gupta2024} underscores the importance of a phased approach to AI development and highlights the significant effort required in the initial stages to ensure a successful implementation. Their approach, however, is purely conceptual, while ours is based on decision-theoretic model.

\citet{shekhar2024optimizing} discuss strategies for optimizing the costs associated with LLM usage. Their research identifies, like ours, key cost variables, such as network costs, embedding costs, and the periodic fine-tuning of models, which are crucial for accurate cost assessment and optimization. They emphasize the importance of considering both fixed and variable costs in the economic evaluation of LLMs. This work complements our study by providing a detailed breakdown of cost components. Additionally, they highlight the impact of reinforcement learning strategies and the necessity of periodically refining models, aligning with our recommendation to integrate more cost variables, including the need for periodic fine-tuning, in future work. Their approach to cost optimization, although, does not takes into account the gains and losses, or different impact of the probabilities of success.

\section{Future work}

Future research should explore more business scenarios to expand the applicability of the proposed models. 
Each industry or business model has unique operational variables and challenges that can affect the adoption and success of LLM. 
By exploring diverse contexts, such as retail, finance, healthcare, and logistics, the framework can be refined to accommodate the specific needs of different sectors. 
This exploration will enhance the robustness of this decision-theoretic approach, ensuring that it is adaptable and relevant across a wide range of business applications.

A comprehensive evaluation of LLMs should consider a broader spectrum of cost variables. Some of these aspects are discussed in \autoref{othercosts}.
This includes not only the initial fixed costs associated with implementation and usage, but also other AI related tasks, such as periodic fine-tuning and maintenance expenses, and additional variable costs, such as network costs. 
Future work should focus on developing models that incorporate these ongoing costs to provide a more accurate and realistic assessment of the total cost of ownership. 

With respect to the models themselves, it is necessary to analyze the impact of the input size on the probability of success with respect to different prompt strategies. 
This would mean modeling $P$, and its variations, a function of $P$, with impact on sensitivity analysis. This would allow for including prompt technical evaluation in our model, since a zero shot prompt in a single turn would be a much smaller transaction than multi-turn mixed strategy with many agents~\citep{sahoo2024systematic,li2024agents}.

A dynamic evaluation of project revenues and costs over time is essential for making informed investment decisions. 
Future work should focus on developing models that calculate the present value of future cash flows, taking into account the value of money over time. 
This involves projecting revenue and cost streams throughout the lifecycle of the project and discounting them to their present value. 
By incorporating techniques such as net present value (NPV) and internal rate of return (IRR), the models can provide a more accurate picture of the long-term financial viability and profitability of LLM investments~\citep{park2013FEE}.

\section{Conclusion}

This study underscores the importance of considering specific business parameters, such as Gain, Losses, and Probability of success, in evaluating the earnings and RoI for LLM. 

This analysis shows that it is important to optimize $P$ and $T$ for the same business operation, when choosing an LLM, always taking into account $C$, although it is fairly small, since all these values have a strong impact on both earnings and RoI.

Similarly, in the second model, the introduction of Probability of being a true positive ($\pvp$) alongside Gain further refines our understanding of earnings and RoI under conditions specific to binary classification tasks. Small values of cost per token show that, for small transaction sizes, LLMs costs could have a very small impact and allow for larger earnings. On the other hand, learning in context, prompt techniques, and multiple agents indicate a trend toward larger transaction size. 

Our findings emphasize that for businesses aiming to optimize their technology investments, particularly in the deployment of advanced language models, it is crucial to consider not only the direct costs, but also the broader economic implications of model performance and success rates. 
We show, for example, that for some business tasks with small transactions and large gains, prompt compression can bring negative financial results.
This approach provides a more nuanced framework that can guide strategic decisions, ensuring that investments are aligned with expected financial returns. 

This paper contributes to the ongoing discourse on computational finance by refining the methods used to assess technology investments, particularly in the field of artificial intelligence.

Future research should consider expanding these analyses to include more complex models and a broader range of operational contexts to further validate and enhance the robustness of our conclusions.


\section*{Acklowdgements}

This paper was written with the help of different artificial intelligence tools to partially generate code, find references, and verify the style and grammar. Among the LLMs used were Google's AI Assistant for Google Colab, Google Search Lab AI, OpenAI ChatGPT-4o, Microsoft Co-Pilot for Visual Studio Code, and Writeful.  All authors have English as their second language. 

The first author is funded by the grant 421793/2022-8 (call CNPq/AWS Nº 64/2022 - Faixa A) in collaboration with Amazon Web Services. This work has been supported by the following Brazilian research agencies: CAPES, and CNpQ. 

This study was financed in part by the Coordenação de Aperfeiçoamento de Pessoal de Nível Superior – Brasil (CAPES) – Finance Code 001”. 
\end{document}